\pdfoutput=1

\documentclass[11pt]{article}

\usepackage{EMNLP2022}

\usepackage{times}
\usepackage{latexsym}

\usepackage[T1]{fontenc}

\usepackage[utf8]{inputenc}

\usepackage{microtype}

\usepackage{inconsolata}

\usepackage{bm}
\usepackage{xcolor}
\usepackage{multirow}
\usepackage{makecell}
\usepackage{textcomp}
\usepackage{graphicx}
\newcommand{\ie}{\textit{i}.\textit{e}., }
\newcommand{\eg}{\textit{e}.\textit{g}., }
\usepackage{amssymb}
\usepackage{amsmath}
\usepackage{comment}
\title{Information-Theoretic Text Hallucination Reduction for Video-grounded Dialogue}
\author{\text{Sunjae Yoon}$^{\dagger}$, \text{Eunseop Yoon}$^{\dagger}$, \text{Hee Suk Yoon}$^{\dagger}$, \text{Junyeong Kim}$^{\ddagger}$, \text{Chang D. Yoo}$^{\dagger}$\Thanks{Corresponding author} \\
         $^{\dagger}$Korea Advanced Institute of Science and Technology (KAIST) \\
         $^{\ddagger}$Chung-Ang University \\ \texttt{\{sunjae.yoon,esyoon97,hskyoon,cd\_yoo\}@kaist.ac.kr}; \texttt{junyeongkim@cau.ac.kr}}

\begin{document}
\maketitle
\begin{abstract}
Video-grounded Dialogue (VGD) aims to decode an answer sentence to a question regarding a given video and dialogue context. Despite the recent success of multi-modal reasoning to generate answer sentences, existing dialogue systems still suffer from a text hallucination problem, which denotes indiscriminate text-copying from input texts without an understanding of the question. This is due to learning spurious correlations from the fact that answer sentences in the dataset usually include the words of input texts, thus the VGD system excessively relies on copying words from input texts by hoping those words to overlap with ground-truth texts. Hence, we design Text Hallucination Mitigating (THAM) framework, which incorporates Text Hallucination Regularization (THR) loss derived from the proposed information-theoretic text hallucination measurement approach. Applying THAM with current dialogue systems validates the effectiveness on VGD benchmarks (i.e., AVSD@DSTC7 and AVSD@DSTC8) and shows enhanced interpretability.
\end{abstract}
%
%
%
\section{Introduction}

Achieving a natural conversational agent that can do ‘look’ (\ie understand what they are seeing) and ‘tell’ (\ie converse what they are thinking) is desiderata in our vision-language community.
%
%
By the broad application of conversational agent, it can potentially assist various subsections of our environment including education, entertainment, security, and visual or other impairments.
For the natural conversation between humans and computers, a video-grounded dialogue (VGD) task \cite{Alamri_2019_CVPR,Hori_2020_DSTC8} has been introduced to generate adequate conversational responses to the queries of humans, while following up on video and dialogue context, which gives more challenging than traditional image-grounded or text-grounded dialogue tasks.
\begin{figure*}[t]
	\centering
	\includegraphics[width=\linewidth]{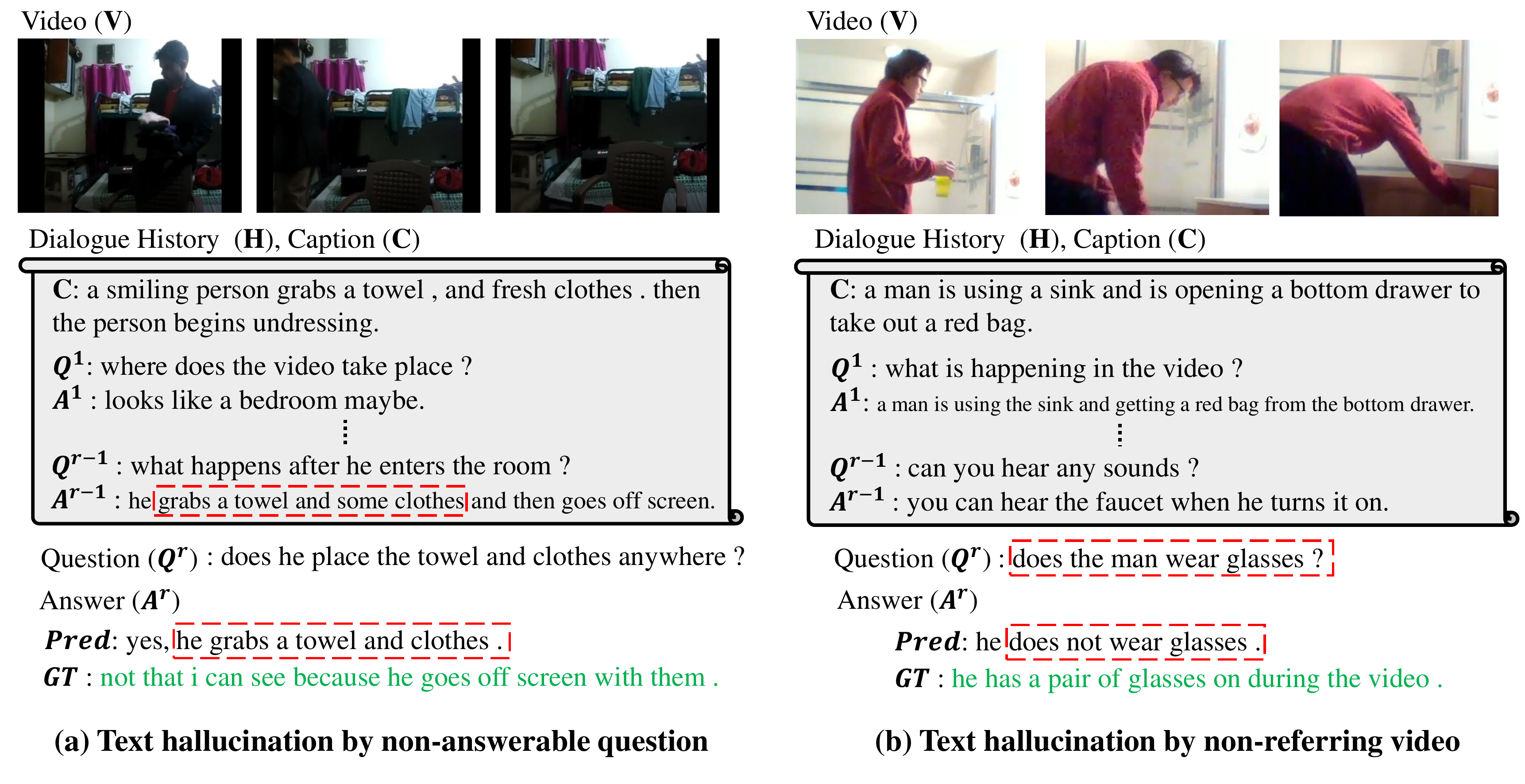}
	\caption{Illustration of video-grounded dialogue system including incorrect answer generation by (a) non-answerable question and (b) non-referring video.}
	\label{fig:1}
\end{figure*}
To be specific, given video $V$, video caption $C$, dialogue history of past Q\&A pairs: $H = \{(Q^{1},A^{1}),...,(Q^{r-1},A^{r-1})\}$, and current $r$-th round question $Q^{r}$, VGD system is expected to make free-form answer sentence $A^{r}$ to given question.
Despite recent advancements in multi-modal interactions including transformer \cite{Vaswani_2017_NIPS}, current VGD systems still suffer text hallucination problem, which denotes indiscriminate text-copying from input texts (\ie question, caption, and dialogue history) to decode answer tokens, but the generated answer sentences are rather inadequate and not related to the question.
This is because current VGD systems learn spurious correlations from the fact that many ground-truth answers in the dataset include partial input texts, thus they perform incorrect text-copy from input texts, namely text hallucination, even in answers where input texts are unnecessary.

Figure \ref{fig:1} gives two indiscriminate text hallucinating cases confounded by spurious correlations in VGD.
As shown in Figure \ref{fig:1}(a), for the given question `does he place the towel and clothes anywhere?', we human identify where the man placed the towel and clothes, and if it cannot be confirmed, we give a sentence meaning `unknown'.
However, in many cases, VGD systems are optimized in situations where they could find clues in video and dialogue, so for a case that they can not find clues, they simply pretend to know the answer by copying texts from input sentences without reasoning why the question is not answerable.
Thus, the VGD systems depend on indiscriminate text hallucination, copying input sentences (\ie questions, caption, dialogue), hoping the copied answer words to overlap the ground-truth words.
Figure \ref{fig:1}(b) presents another dependence on this text hallucination even in the answerable question.
Given the question of `does the man wear glasses?', the current VGD system provides incorrect answer without referring to the video and focuses on pretending to know the answer via copying input texts.
This is because the system is holding overconfidence in the text hallucination, such that it ignores the meaning of the question and video.
Therefore, current VGD systems are prone to rely on language model tainted with incorrect text hallucination, which hinders them from accurately learning question-answer association.
\begin{figure}[t]
	\centering
	\includegraphics[width=\linewidth]{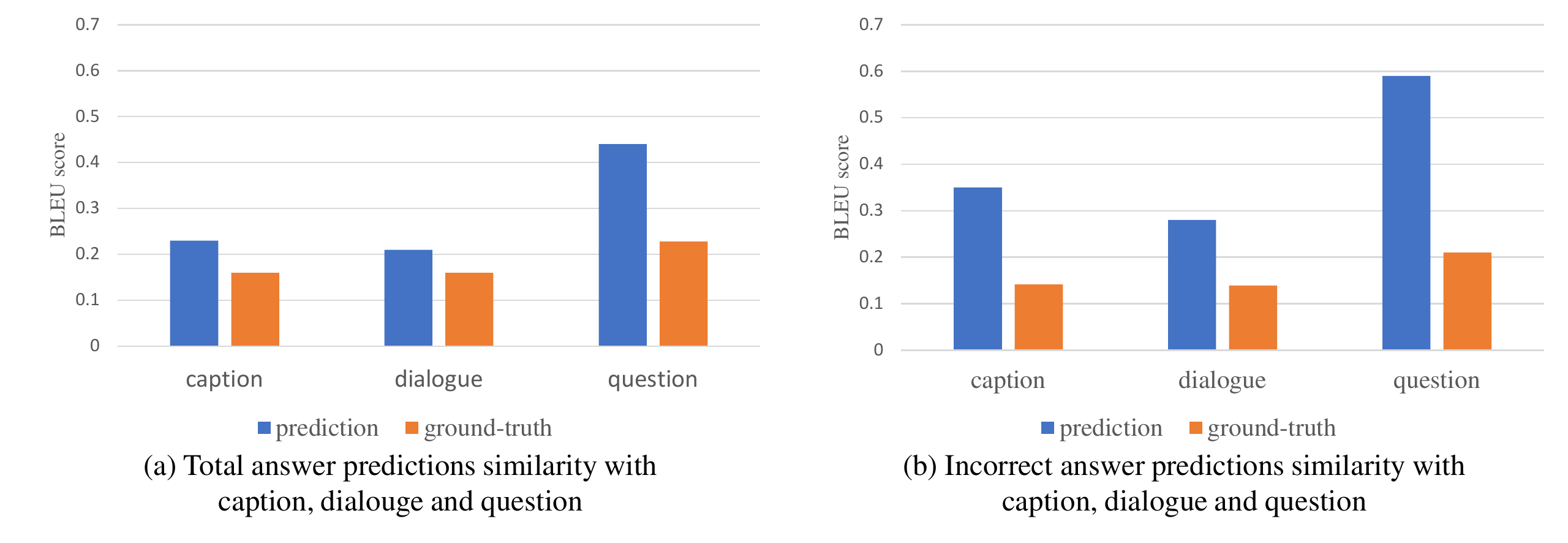}
	\caption{(a) sentence similarity scores (BLEU score) between input sentences (caption, dialogue, question) and answer sentence (prediction, ground-truth). (b) sentence similarity between input sentences and answer sentences (incorrect predictions, ground-truth), which tells that incorrect answers have made mistakes by hallucinating input sentences.}
	\label{fig:2}
\end{figure}

Our manual studies in Figure \ref{fig:2} give experimental evidence that the answer sentences predicted by current VGD systems \cite{Le_2019_ACL,Li_2021_tip} are dependent on indiscriminate text hallucination.
Figure \ref{fig:2}(a) presents sentence similarity score, BLEU \cite{papineni2002bleu}, which computes word overlapping between (1) predicted answers and input texts (\ie caption, dialogue and question), and (2) ground-truth answers and input texts from AVSD\footnote{Audio-Visual Scene Aware Dialog \cite{Alamri_2019_CVPR}} validation dataset.
The higher scores between predicted answers and input texts explain the reliance on input texts for decoding answer tokens.
We may take this for granted, but as shown in Figure \ref{fig:2}(b), the problem gets distinguishable when collecting all the `incorrect'\footnote{Here, we regard predictions with a BLEU score of less than 0.1 as `incorrect'.} predictions.
Many failure cases (\ie incorrect predictions) include that the predicted answers are more similar to input texts, which proves indiscriminate text hallucination without the understanding of given questions and videos.
%
%
%

One straightforward solution to mitigate this indiscriminate text hallucination is to extend the dataset using augmentations or modulating answer descriptions to be more stereoscopic.
However, the augmentation has limitations in terms of diversity and the modulated descriptions can be sometimes ad-hoc and unnecessarily extravagant.
%
%
Intrigued by the current overconfidence in text hallucination of VGD systems, we contrive to build Text Hallucination Mitigating (THAM) framework that mitigates feature-level hallucination effects via introducing information-theoretic regularization.
THAM framework incorporates Text Hallucination Regularization (THR) loss derived from the mutual information between the response language model and the proposed hallucination language model.
Minimizing THR loss contributes to reducing indiscriminate text copying and boosting dialogue performances.
%
%
%
%
%
THAM validates effectiveness with steady performance gain on top of the current several runner models \cite{Hori_2019_ICASSP,Le_2019_ACL,kim_2021_AAAI,Li_2021_tip} via a model-agnostic approach.
experimental results show state-of-the-art performances on two VGD benchmarks (\ie AVSD@DSTC7 and AVSD@DSTC8) and enhanced interpretability.
\section{Related Work}
\subsection{Video-grounded Dialogues}
\label{sec:2.1}
Visual Question Answering (VQA) \cite{Antol_2015_ICCV,li2022invariant,xiao2022video} is one of the proxy tasks for evaluating multi-modal understanding of vision-language systems.
The recent success of natural language processing \cite{devlin2018bert,radford2019language} gives a bridge to advance VQA for video-grounded dialogue (VGD) system \cite{Alamri_2019_CVPR,Hori_2020_DSTC8}, which aims to generate open-ended answer sentence founded on video and dialogue of human.
For this VGD, many recurrent neural networks \cite{Nguyen_2019_DSTC7,Sanabria_2019_DSTC7} have been proposed to hold meaningful semantics along the consecutive dialogues, and a transformer-based VGD system \cite{Li_2021_tip} has also been introduced to enhance multi-modal interaction between video and text, including word-embedding attention \cite{Lee_2020_DSTC8}, hierarchical attention \cite{Le_2019_DSTC7} and pointer-augmented decoding \cite{Le_2020_DSTC8}.
Furthermore, graph representation is considered to connect common semantics among intra-frames and inter-frames \cite{geng_2021_AAAI} and to uncover co-referencing between frames and texts \cite{kim_2021_AAAI}.
However, these systems still suffer from the hallucination problem in generating answer sentences and for this problem, we proposed an information-theoretic text hallucination mitigating framework.
\section{Preliminaries}
\subsection{Estimating Mutual Information}
\label{sec:3.1}
To identify the feature-level text hallucination, we first introduce the mutual information $I(\cdot ; \cdot)$, which measures co-dependence between two random variables $X$ and $Y$ over the space $\mathcal{X} \times \mathcal{Y}$ like below:
\begin{eqnarray}
I(X;Y) := H(X) - H(X|Y),
\end{eqnarray}
where $H(\cdot)$ is the Shannon entropy and $H(X|Y)$ is the conditional entropy of $X$ given $Y$.
This mutual information is also equal to the Kullback-Leibler (KL-) divergence $D_{KL}(\cdot || \cdot)$ between joint probability distribution $P_{XY}$ and the product of marginals $P_{X} \otimes P_{Y}$ like below:
\begin{eqnarray}
I(X;Y) = D_{KL}(P_{XY} || P_{X} \otimes P_{Y}),
\end{eqnarray}
where, given two probability distributions $p(x)$ and $q(x)$ on variable $x$, KL divergence is defined as:
\begin{eqnarray}
D_{KL}(p || q) := \mathbb{E}_{x \sim p}[\textrm{log}(\frac{p(x)}{q(x)})].
\end{eqnarray}
%
As the KL divergence increases, the co-dependence between $X$ and $Y$ becomes stronger.
However, calculating KL divergence is tractable for only a few cases (\ie discrete variables), as it is unavailable to hold exact distributions of the training dataset.
Recent approach \cite{belghazi2018mine} is performed on estimating mutual information for continuous high-dimensional variables using neural network founded on the Donsker-Varadhan representation\footnote{It provides a supremum of the KL divergence over all functions $T: D_{KL}(P||Q) = \textrm{sup}_{T:\mathbb{R}^{D} \rightarrow \mathbb{R}}\mathbb{E}_{P}[T] - log(\mathbb{E}_{Q}[\textrm{exp}(T)])$.} \cite{donsker1975asymptotic} defined below:
\begin{align}
\begin{split}
I(X;Y) &\geq I_{\phi}(X;Y) \\
= \textrm{sup}_{\phi \in \Phi} \mathbb{E}_{P_{XY}}[T_{\phi}] &- \textrm{log}(\mathbb{E}_{P_{X}\otimes P_{Y}}[e^{T_{\phi}}]),
\end{split}
\label{eq:4}
\end{align}
where $T_{\phi}: \mathbb{R}^{D} \rightarrow \mathbb{R}$ is a neural network parameterized by $\phi \in  \Phi$, and the expectations of $\mathbb{E}_{{P}_{XY}}$ and $\mathbb{E}_{{P}_{X}\otimes {P}_{Y}}$ are approximated by empirical sampling.
Thus, maximizing $I_{\phi}(X;Y)$ provides a tight lower bound of original mutual information\footnote{Refer proofs in the appendices.} $I(X;Y)$.
\subsection{Video-grounded Dialogue Task}
Video-grounded Dialogue (VGD) aims to produce free-form natural language answer for a given question. 
In the formal definition of the VGD task \cite{Alamri_2019_CVPR}, VGD system takes tuples $(v,h,q^{r})$ as inputs and produces answer sentence $a^{r}$, where $v$ is video, $h$ is dialogue history and $q_{r}$ question asked at current round $r \in \{1,\cdots,R\}$.
Here, the dialogue history $h = \{c,(q^{1},a^{1}),\cdots,(q^{r-1},a^{r-1})\}$ is a set of question-answer pairs of previous rounds and caption $c$ describing the summary of the video.
For training of the VGD system, we perform next-word prediction, where it is trained to predict $t$-th answer word token $a^{r}_{t}$ for given inputs of tuples ($v$,$h$,$q^{r}$) and partial answer word tokens $a^{r}_{<t}$ before $t$-th.
\begin{figure*}[t]
	\centering
	\includegraphics[width=\linewidth]{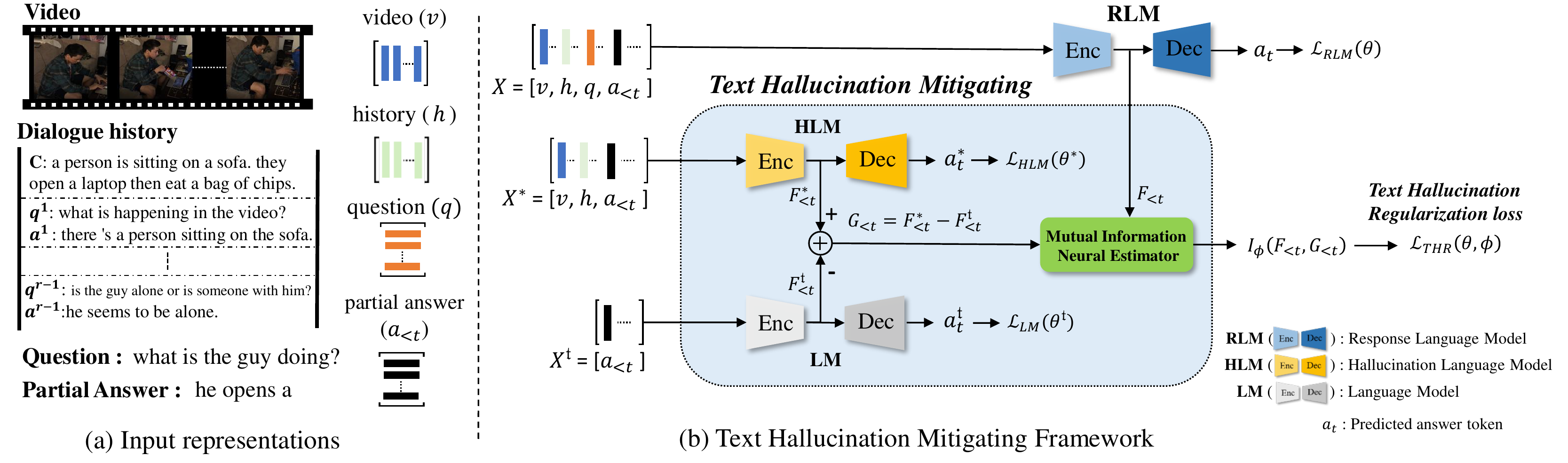}
	\caption{Illustration of Text Hallucination Mitigating Framework (THAM) for video-grounded dialogue. THAM mitigates feature-level hallucination effects in Response Language Model via introducing Text Hallucination Regularization (THR) loss, where THR aims to minimize mutual information between encoder features of RLM and features from Hallucination Language Model.}
	\label{fig:3}
\end{figure*}
\section{Text Hallucination Mitigating Framework}
\label{sec:4.1}
In Figure \ref{fig:3}, to build Text Hallucination Mitigating (THAM) framework, we prepare three different language models composed of encoder-decoder pairs: (1) Response Language Model (RLM), (2) Hallucination Language Model (HLM), and (3) Language Model (LM).
RLM is a naive VGD model, such that it is given complete samples of $v$, $h$, $q$, and partial answer $a^{r}_{<t}$ to predict the next answer token $a^{r}_{t}$.
HLM is designed to generate answer tokens relying on the text hallucination, where HLM is given deficient input texts (\ie $h,a^{r}_{<t}$) without question, which is unavailable to reason the correct answer and inevitably relies on hallucinating sentence to overlap with ground-truth words via copying input texts without knowledge of the question.
Using this HLM, our proposed Text Hallucination Regularization (THR) mitigates feature-level hallucination effects in the RLM via minimizing the mutual information between the features of RLM encoder and hallucinating features of HLM encoder.
%
%
However, not all the features of HLM are bad, because HLM, as a language model, is also trained to make a grammatically complete sentence, where those grammatical knowledge should be removed before performing THR. 
Therefore we train another language model (LM), which predicts the next answer token $a_{t}^{r}$ from only given partial answer $a_{<t}^{r}$.
We remove encoder features of LM from those of HLM in advance and apply the THR loss.
\subsection{Input representations}
We give formal feature definitions of $v$, $h$, $q^{r}$ and $a^{r}$ embedded into $d$-dimensional space.
Following \cite{hori2019end,Li_2021_tip}, for the video features, we utilize the I3D model \cite{carreira2017quo} pre-trained on YouTube videos and the Kinetics dataset \cite{kay2017kinetics} to get 2048-dimensional rgb features $\mathbf{v}_{rgb} \in \mathbb{R}^{L \times 2048}$ and optical flow features $\mathbf{v}_{opt} \in \mathbb{R}^{L \times 2048}$ in the images, where the $L$ is the number of video frames.
Audio features are also available in the video of the AVSD dataset, we get 128-dimensional features\footnote{Interpolation is considered for audio features to be synchronized with video features.} $\mathbf{v}_{aud} \in \mathbb{R}^{L \times 128}$ using pre-trained VGGish \cite{hershey2017cnn}.
The aforementioned three features are concatenated along feature dimension axis and embedded into $d$-dimensional space as:
\begin{eqnarray}
\mathbf{v} = [\mathbf{v}_{rgb}||\mathbf{v}_{opt}||\mathbf{v}_{aud}]W_{\mathbf{v}} \in \mathbb{R}^{L \times d},
\end{eqnarray}
where $W_{\mathbf{v}} \in \mathbb{R}^{(2048+2048+128) \times d}$ is $d$-dimensional embbeder and $[\cdot||\cdot]$ denotes concatenation.

For the text features, we follow the T5-base Transformer \cite{raffel2020exploring} and tokenize all the sentences (\ie $q^{r}, h, a^{r}$) into a series of WordPieces \cite{wu2016google}. 
The final representations for each sub-word token are obtained by summing up their token embeddings and relative positional embeddings, followed by a layer normalization \cite{Ba_2016_arxiv}.
We give formal definitions of them as: history $\mathbf{h} \in \mathbb{R}^{L_{\mathbf{h}} \times d}$, question $\mathbf{q} \in \mathbb{R}^{L_{\mathbf{q}} \times d}$ and answer $\mathbf{a} \in \mathbb{R}^{L_{\mathbf{a}} \times d}$, where $L_{\mathbf{h}}, L_{\mathbf{q}}$ and $L_{\mathbf{a}}$ are the number of tokens of each text\footnote{We delete superscript $r$ in the notations of features for simplicity.}.
\subsection{Text Hallucination Regularization}
Text Hallucination Regularization (THR) is designed for the VGD model (\ie RLM) to mitigate indiscriminate text hallucination (\ie text or word copying) from input texts without understanding of the question.
As we describe the methodology of THAM in Section \ref{sec:4.1}, here, we focused on mathematical formulations for the reproducibility of THAM with proposed THR.
%
%
%
\paragraph{Training language models.}
To prepare own purpose of three language models (\ie RLM, HLM, LM), as the first stage, we train them with their defined inputs in the followings.
Response Language Model (RLM) is designed for original purpose of VGD, where it is given complete input sample as $X_{<t} = [\mathbf{v}||\mathbf{h}||\mathbf{q}||\mathbf{a}_{<t}]$ and trained to generate next word tokens for answer sentence $a^{r} = \{ a^{r}_{1},\cdots,a^{r}_{m}\}$ with sentence length $m$ using cross-entropy loss like below:
\begin{eqnarray}
\mathcal{L}_{RLM}(\theta) = \textrm{log}\prod_{t=1}^{m} P(a^{r}_{t}|X_{<t};\theta).
\end{eqnarray}
Hallucination Language Model (HLM) is intended to learn reliance on text hallucination effects for generating an answer. 
To train HLM, we utilize the fact that ground-truth answer sentences of VGD are usually similar to the partial texts of inputs.
Therefore, we give the HLM with deficient input texts $X^{\star}_{<t} = [\mathbf{v}||\mathbf{h}||\mathbf{a}_{<t}]$ without question like:
\begin{eqnarray}
\mathcal{L}_{HLM}(\theta^{\star}) = \textrm{log}\prod_{t=1}^{m} P(a^{r}_{t}|X^{\star}_{<t};\theta^{\star}),
\end{eqnarray}
where the deficient input texts make it difficult for HLM to perform correct answer reasoning. 
(See more results in the ablation studies of Table \ref{tab:3}.)
In the optimization, although the HLM can identify the similarities between partial texts of inputs and ground-truth answers, but it is unavailable to learn why the answers are similar to input texts, which results in training of the text hallucination.
%
Using this overconfidence in text hallucination of HLM, we build Text Hallucination Regularization (THR) loss to mitigate the text hallucinating effect in naive RLM in the following.
\paragraph{Text Hallucination Regularization.}
Text Hallucination Regularization (THR) is introduced to mitigate indiscriminate text hallucination of VGD models to answer the question.
THR loss is defined by feature-level mutual information between RLM and HLM.
To this, we first define encoder features of each trained model: (1) RLM's encoder features as $F_{<t} = f_{RLM}(X_{<t},\theta) \in \mathbb{R}^{d}$ and (2) HLM's encoder features as  $F^{\star}_{<t} = f_{HLM}(X^{\star}_{<t},\theta^{\star}) \in \mathbb{R}^{d}$, where $f$ denotes the transformer encoders of each model. 
%
%
These two features (\ie $F_{<t}, F^{\star}_{<t}$) are outputs from the position of $\mathbf{a}_{t-1}$ in the transformer.
Here, we refer to $F_{<t}$ as `factual' features and $F^{\star}_{<t}$ as `hallucinating' features.
Our proposed THR aims to hold feature-level independence between factual features and hallucinating features via minimizing mutual information among them. 
However the grammatical knowledge in $F^{\star}_{<t}$ to build language sentence still should be correlated with $F_{<t}$, as both language models are trained from grammatically complete ground-truth language sentences.
Thus, we prepare pure language model (LM), which predicts answer token with only given partial answer tokens $X^{\dagger}_{<t} = [\mathbf{a}_{<t}]$:
\begin{eqnarray}
\mathcal{L}_{LM}(\theta^{\dagger}) = \textrm{log}\prod_{t=1}^{m} P(a^{r}_{t}|X^{\dagger}_{<t};\theta^{\dagger}),
\end{eqnarray}
where we get pure language features $F_{<t}^{\dagger} = f_{LM}(X_{<t}^{\dagger},\theta^{\dagger}) \in \mathbb{R}^{d}$ from the LM's encoder, which has the only grammatical knowledge to make complete language. 
We remain pure hallucinating effects via subtracting the language features $F_{<t}^{\dagger}$ from the hallucinating features $F^{\star}_{<t}$:
\begin{eqnarray}
G_{<t} = F_{<t}^{\star} - F_{<t}^{\dagger} \in \mathbb{R}^{d},
\end{eqnarray}
where the $G_{<t}$ is the pure hallucinating (pure-h) features, which hold hallucinating effects without grammatical knowledge.
%
%
Founded on factual features $F_{<t}$ and pure-h features $G_{<t}$, we finally define THR loss. 
THR loss calculates feature-level mutual information between  $F_{<t}$ and $G_{<t}$.
Thanks to the mutual information neural estimator (MINE) \cite{belghazi2018mine}, we get high-dimensional mutual information between the $F_{<t}$ and the $G_{<t}$, where we utilize it as THR loss for a regularization:
\begin{eqnarray}
\mathcal{L}_{THR}(\theta,\phi) = I_{\phi}(f_{RLM}(X_{<t},\theta); G_{<t})
\end{eqnarray}
By minimizing $\mathcal{L}_{THR}(\theta,\phi)$ with respect to the parameter $\theta$, we train the RLM to be independent of HLM's indiscriminate text hallucination\footnote{The $\theta^{\star}$, $\theta^{\dagger}$ in $G_{<t}$ are different parameters with $\theta$ in $F_{<t}$.}.
Following the maximizing lower bound of estimated mutual information in Equation \ref{eq:4}, the final objective function is formulated as:
\begin{eqnarray}
\underset{\theta}{\textrm{min}} \hspace{0.05cm} \underset{\phi}{\textrm{max}}\hspace{0.05cm} \mathcal{L}_{RLM}(\theta) +  \alpha \mathcal{L}_{THR}(\theta,\phi)
\end{eqnarray}
where $\alpha$ is a hyperparameter and the objective function is a minimax problem, we alternate to train and update the parameters $\theta$ and $\phi$ in every epoch.
%

%
%
%
%
%
%
\begin{table*}[t]
	\centering
	\caption{Experimental results on the test split of AVSD benchmark at DSTC7 and DSTC8 challenges (B: BELU, M: METEOR, R: ROUGE-L, C: CIDEr, cp: caption, $\star$: reported in \cite{kim_2021_AAAI}).}
	\begin{tabular}{l||c c c c c c c}
		\Xhline{3\arrayrulewidth}
		\multicolumn{8}{c}{AVSD@DSTC7}\\ \Xhline{3\arrayrulewidth}
		Methods                              & B1     & B2     &B3     & B4     & M   & R   & C     \\ 
		\Xhline{2\arrayrulewidth}
		Baseline \cite{Hori_2019_ICASSP}     & 0.621     & 0.480     & 0.379     & 0.305     & 0.217    & 0.481     & 0.733     \\
		HMA \cite{Le_2019_DSTC7}             & 0.633     & 0.490     & 0.386     & 0.310     & 0.242    & 0.515     & 0.856     \\
		RMFF \cite{Yeh_2019_DSTC7}           & 0.636     & 0.510     & 0.417     & 0.345     & 0.224    & 0.505     & 0.877     \\
		EE-DMN \cite{Lin_2019_DSTC7}         & 0.641     & 0.493     & 0.388     & 0.310     & 0.241    & 0.527     & 0.912     \\
		JMAN \cite{Chu_2020_DSTC8}           & 0.667     & 0.521     & 0.413     & 0.334     & 0.239    & 0.533     & 0.941     \\
		FA-HRED \cite{Nguyen_2019_DSTC7}     & 0.695     & 0.553     & 0.444     & 0.360     & 0.249    & 0.544     & 0.997     \\
		CMU \cite{Sanabria_2019_DSTC7}       & 0.718     & 0.584     & 0.478     & 0.394     & 0.267    & 0.563     & 1.094     \\
		MSTN \cite{Lee_2020_DSTC8}           & -         & -         & -         & 0.377     & 0.275    & 0.566     & 1.115     \\
		JSTL \cite{Hori_2019_Interspeech} w/o cp    & 0.675     & 0.543     & 0.446     & 0.371     & 0.248    & 0.527     & 0.966  \\
		JSTL \cite{Hori_2019_Interspeech}    & 0.727     & 0.593     & 0.488     & 0.405     & 0.273    & 0.566     & 1.118     \\
		MTN$^{\star}$ \cite{Le_2019_ACL}               & 0.731     & 0.597     & 0.490     & 0.406     & 0.271    & 0.564     & 1.127     \\
		MTN-P \cite{Le_2020_DSTC8}           & 0.750& 0.619     & 0.514     & 0.427     & 0.280    & 0.580& 1.189     \\
		VGNMN \cite{le2022vgnmn}                     & -     & -     & -     & 0.429     & 0.278    & 0.578     & 1.188     \\
		SCGA \cite{kim_2021_AAAI}                                 & 0.745     & 0.622& 0.517& 0.430&0.285& 0.578     & 1.201\\	
		RLM \cite{Li_2021_tip} &	0.765     & 0.643 & 0.543 & 0.459 &0.294 &0.606 &1.308\\	\hline
		\bf{T5RLM (Ours)}                                  & 0.767 &0.644 &0.542 &0.461 &0.296 &0.608 &1.311\\
		\textbf{THAM (T5RLM + THR loss)}                                 & \bf{0.778} &\bf{0.654} &\bf{0.549} &\bf{0.468} &\bf{0.308} &\bf{0.619} &\bf{1.335}\\	
		\Xhline{3\arrayrulewidth} 
		\multicolumn{8}{c}{AVSD@DSTC8}\\ \Xhline{2\arrayrulewidth}
		MDMN \cite{Xie_2020_DSTC8}           & -       & -       & -       & 0.296   & 0.214  & 0.496   & 0.761 \\
		JMAN \cite{Chu_2020_DSTC8}           & 0.645   & 0.504   & 0.402   & 0.324   & 0.232  & 0.521   & 0.875 \\
		STSGR \cite{Geng_2020_DSTC8}         & -       & -       & -       & 0.357   & 0.267  & 0.553   & 1.004 \\
		MSTN \cite{Lee_2020_DSTC8}           & -       & -       & -       & 0.385   & 0.270  & 0.564   & 1.073 \\
		MTN-P \cite{Le_2020_DSTC8}           & 0.701   & 0.587   & 0.494   & 0.419   & 0.263  & 0.564   & 1.097 \\
		SCGA \cite{kim_2021_AAAI} w/o cp                     & 0.675   & 0.559   & 0.459   & 0.377   & 0.269  & 0.555   & 1.024  \\
		SCGA \cite{kim_2021_AAAI}                                 & 0.711   & 0.593   & 0.497 & 0.416   & 0.276  & 0.566   & 1.123 \\ 
		RLM \cite{Li_2021_tip}                                 & 0.746     & 0.626 & 0.528 & 0.445 &0.286 &0.598     & 1.240\\ \hline
		\bf{T5RLM (Ours)}                             & 0.749 &0.631 &0.529 &0.445 &0.290 &0.600 &1.263\\
		\bf{THAM (T5RLM + THR loss)}                                 &\bf{0.764}     &\bf{0.641} &\bf{0.538} &\bf{0.455} &\bf{0.301} &\bf{0.610} &\bf{1.304}\\	
		\Xhline{3\arrayrulewidth}
	\end{tabular}
	\label{tab:1}
\end{table*}

\begin{table*}[t]
	\centering
	\caption{Experimental results on the test split of AVSD benchmark at DSTC7 and DSTC8 challenges for applying THR loss on VGD runner models (B1 = BELU1, $*$: reconstruction-based results, $\dagger$: single reference results).}
	\begin{tabular}{l||c c c c c c c}
		\Xhline{3\arrayrulewidth}
		\multicolumn{8}{c}{AVSD@DSTC7}\\ \Xhline{3\arrayrulewidth}
		Methods                              & B1     & B2     &B3     & B4     & METEOR   & ROUGE-L   & CIDEr     \\ 
		\Xhline{2\arrayrulewidth}
		Baseline \cite{Hori_2019_ICASSP}                    & 0.621 & 0.480 & 0.379 & 0.305 & 0.217  & 0.481 & 0.733 \\
		\bf{Baseline + THR loss}                            & 0.635 & 0.495 & 0.388 & 0.313 & 0.230  & 0.492 & 0.762     \\\hline
		$\textrm{MTN}^{\dagger}$ \cite{Le_2019_ACL}         & 0.357 & 0.241 & 0.173 & 0.128 & 0.162  & 0.355 & 1.249     \\
		\bf{$\textrm{MTN}^{\dagger}$ + THR loss}            & 0.371 & 0.252 & 0.181 & 0.136 & 0.175  & 0.374 & 1.265     \\ \hline
		$\textrm{SCGA}^{\star}$ \cite{kim_2021_AAAI}        & 0.746 & 0.618 & 0.514 & 0.428 & 0.283  & 0.575 & 1.193\\	
		\bf{$\textrm{SCGA}^{\star}$ + THR loss}             & 0.758 & 0.629 & 0.522 & 0.430 & 0.295  & 0.587 & 1.214\\	\hline
		RLM \cite{Li_2021_tip}                              & 0.765 & 0.643 & 0.543 & 0.459 & 0.294  & 0.606 & 1.308\\
		\bf{RLM + THR loss}                                 & 0.775 & 0.651 & \bf{0.551} & 0.465 & 0.305 & 0.616 & 1.331\\\hline
		\bf{T5RLM (Ours)}                                   & 0.767 &0.644  &0.542 &0.461 &0.296 &0.608 &1.311\\	
		\bf{T5RLM + THR loss}                               & \bf{0.778} &\bf{0.654} &0.549 &\bf{0.468} &\bf{0.308} &\bf{0.619} &\bf{1.335}\\ \Xhline{3\arrayrulewidth}
		\multicolumn{8}{c}{AVSD@DSTC8}\\ \Xhline{3\arrayrulewidth}
		$\textrm{MTN}^{\star}$ \cite{Le_2019_ACL}          & 0.691   & 0.570   & 0.471   & 0.402   & 0.252  & 0.549   & 1.043 \\
		\bf{$\textrm{MTN}^{\star}$ + THR loss}          & 0.707   & 0.582   & 0.481   & 0.409   & 0.265  & 0.563   & 1.079 \\ \hline
		$\textrm{SCGA}^{\star}$ \cite{kim_2021_AAAI}                                 & 0.706 & 0.587   & 0.498 & 0.412   & 0.277  & 0.563   &1.113 \\
		\bf{$\textrm{SCGA}^{\star}$  + THR loss}                               & 0.727 & 0.603   & 0.507 & 0.425   & 0.289  & 0.581   & 1.169 \\ \hline
		RLM \cite{Li_2021_tip}                                 & 0.746     & 0.626 & 0.528 & 0.445 &0.286 &0.598     & 1.240\\
		\bf{RLM + THR loss}                                 & 0.762     & 0.639 & 0.537 & 0.452 &0.299 &0.607     & 1.287\\ \hline
		\bf{T5RLM (Ours)}                                  & 0.749 &0.631 &0.529 &0.445 &0.290 &0.600 &1.263\\
		\bf{T5RLM + THR loss}                                  &\bf{0.764}     &\bf{0.641} &\bf{0.538} &\bf{0.455} &\bf{0.301} &\bf{0.610} &\bf{1.304}\\	
		\Xhline{3\arrayrulewidth}
	\end{tabular}
	\label{tab:2}
\end{table*}

\section{Experiments}
\subsection{Datasets}
\paragraph{AVSD@DSTC7 and AVSD@DSTC8.} (Audio-Visual Scene Aware Dialog) \cite{Alamri_2019_CVPR,Hori_2020_DSTC8} is a popular benchmark dataset for VGD, where each dialogue includes 10 pairs of question and answer for one video.
The video is collected from Charades \cite{Sigurdsson_2016_ECCV} human-activity dataset and has a short description summarizing overall scenes in the video.
AVSD@DSTC 7 and 8 are released for Dialogue System Technology Challenge (DSTC), where AVSD@DSTC7 contains $7,659$, $1,787$, and $1,710$ dialogues for training, validation and test, but AVSD@DSTC8 is only provided with $1,710$ dialogues for test in the second challenge. 
For test-set evaluation, $6$ reference answers are provided.
\subsection{Metrics}
We follow official natural language generation metrics for AVSD benchmark (\ie BLEU, METEOR \cite{banerjee2005meteor}, ROUGE-L \cite{lin2004rouge}, and CIDEr \cite{vedantam2015cider}). The metrics are provided by challenge organizers\footnote{\textrm{github.com/dialogtekgeek/DSTC8-AVSD\_official}} and formulated to compute the word overlapping between each generated answer and reference answer.
\subsection{Results on AVSD benchmark}
Table \ref{tab:1} summarizes the experimental results on the AVSD dataset.
THAM is compared to several previous results of VGD systems (Please refer the descriptions about these VGD systems in the sec \ref{sec:2.1} of the Related Work.), where the performances of the official six references are evaluated on AVSD@DSTC7 and AVSD@DSTC8.
\begin{table}[t]
	\centering
	\caption{Ablation study on variants of HLM to learn indiscriminate text hallucination from different text inputs on the valid split of AVSD@DSTC7. (single reference)}
	\begin{tabular}{l|| c c}
		\Xhline{3\arrayrulewidth}
		 Input variants on HLM                           & BELU1 & CIDEr \\ 
		\Xhline{2\arrayrulewidth}
		 $X_{<t}^{\star} = [\mathbf{h}||\mathbf{a}_{<t}]$                                & 0.324 & 1.513 \\ \hline
		 \ \ $X_{<t}^{\star} = [\mathbf{q}||\mathbf{a}_{<t}]$           & 0.289 & 1.329 \\
		 \ \ $X_{<t}^{\star} = [\mathbf{v}||\mathbf{a}_{<t}]$           & 0.275 & 1.215 \\
		\ \ $X_{<t}^{\star} = [\mathbf{v}||\mathbf{h}||\mathbf{a}_{<t}]$           & 0.309 & 1.482 \\
		 \ \ $X_{<t}^{\star} = [\mathbf{h}||\mathbf{q}||\mathbf{a}_{<t}]$           & 0.279 & 1.306 \\
		\Xhline{3\arrayrulewidth}
	\end{tabular}
	\label{tab:3}
\end{table}
To validate the effectiveness of proposed our THR loss, we report performances of our naive VGD model (\ie RLM) based on the T5 Transformer \cite{raffel2020exploring}. 
Here, we use `T5RLM' for the terminology of our RLM to avoid confusion with RLM in \cite{Li_2021_tip} based on GPT2 Transformer \cite{radford2019language}.
\begin{figure}[t]
	\centering
	\includegraphics[width=\linewidth]{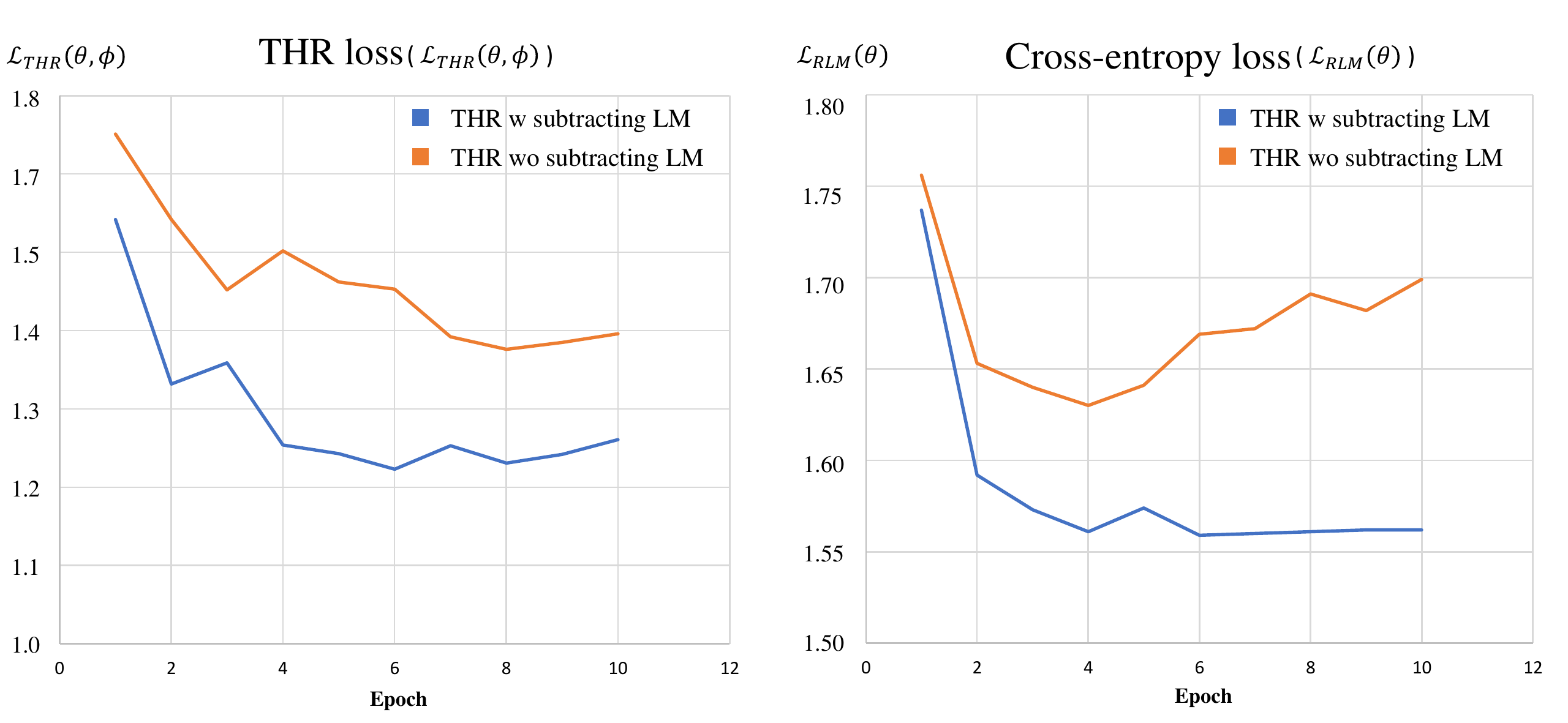}
	\caption{Illustration of THR loss (left) and cross-entropy loss (right) along the epoch on valid split of AVSD@DSTC7 with and without subtracting the encoder features of LM from the encoder features of HLM}
	\label{fig:4}
\end{figure}
In the method, we select a Transformer-base encoder for THAM for its simplicity. 
However, as our framework can be applied to any other VGD systems in a model-agnostic manner, we also validate its effectiveness on recent runner VGD models in Table \ref{tab:2}.
\begin{figure}[t]
	\centering
	\includegraphics[width=\linewidth]{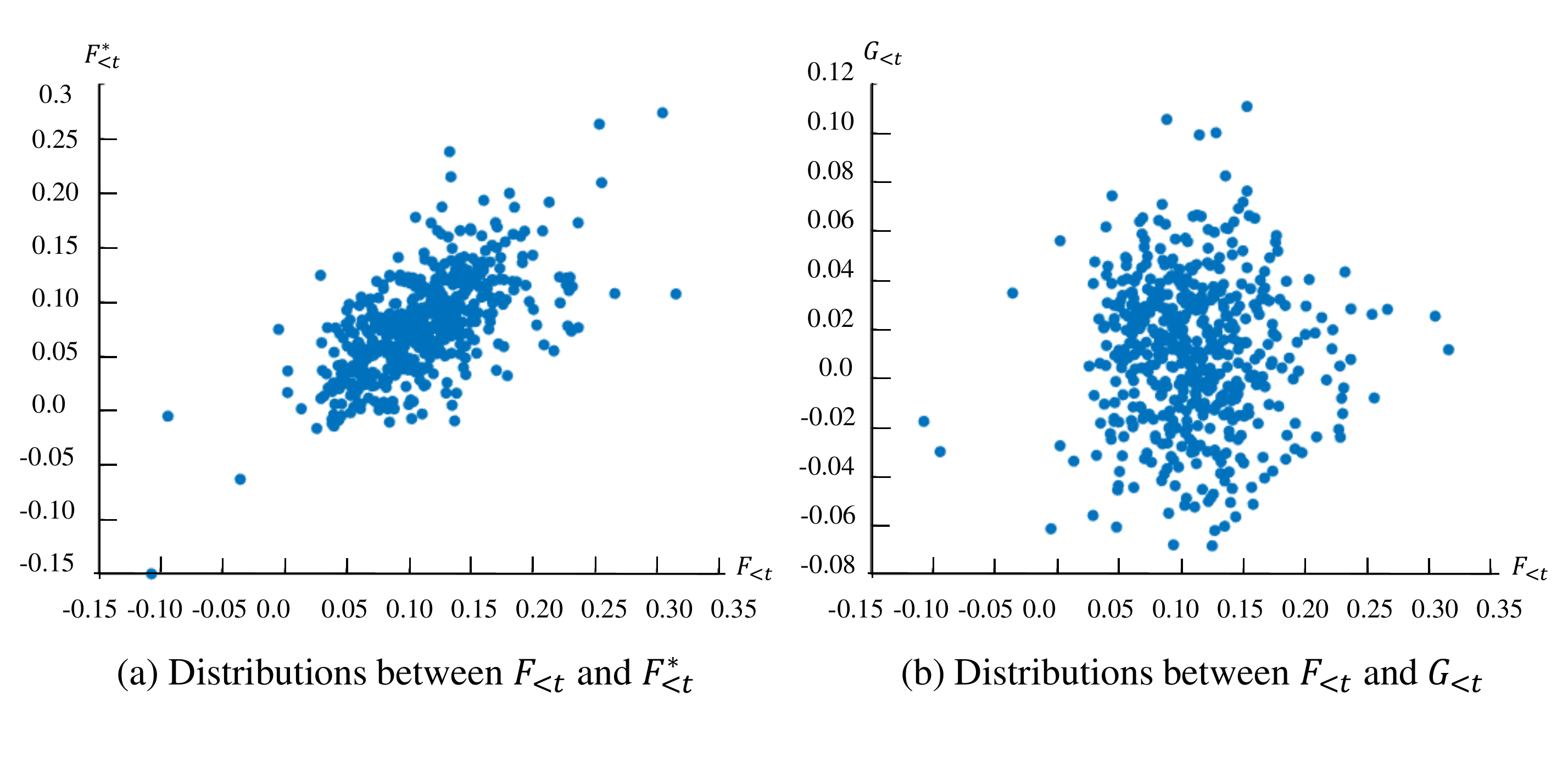}
	\caption{Joint distributions of encoder features between (a) RLM ($F_{<}t$) and HLM ($F_{<t}^{\star}$), (b) RLM ($F_{<}t$) and HLM with subtracting LM ($G_{<}t$). (a) shows correlations with $F_{<t}^{\star}$ by grammatical knowledge in HLM, and (b) shows relatively independent distributions by $G_{<t}$.}
	\label{fig:5}
\end{figure}
In detail, we reproduce the MTN, SCGA and RLM from their public papers and codes.
For the MTN, we measure predicted answers with a single reference following the original work of it.
On top of VGD models, THR loss show steady performance gain on both AVSD datasets.
\subsection{Ablation Study}
Table \ref{tab:3} summarizes the THAM results on input variants of HLM.
HLM is designed to build excessive text conjugating language models via giving inputs that can not infer the correct answer.
In the optimization, it is just optimized to learn spurious correlations between inputs $X_{<t}^{\star}$ and outputs $a^{r}$.
Introducing history only for the inputs of HLM shows the most effectiveness.
We consider this is because the history (\ie dialogue history) contains a relatively large amount of texts, but without question, it is just captions that can not infer the answer.
%
Here, the HLM inevitably learns indiscriminate text hallucination as HLM does not know the question: text hallucination as a result of copying a sentence from input sentences can lead to greater overlap with the ground-truth answer than simply generating an answer without knowing the question.
Conversely, we also devise the HLM with an input of question without history, which was not effective in THAM performance. 
We consider that this is because the AVSD dataset includes some samples, where the correct answer can be easily inferred from a question alone without any other modalities, thus text conjugating on the question should be beneficial.

Figure \ref{fig:4} shows THR loss $\mathcal{L}_{THR}(\theta,\phi)$ and cross-entropy loss $\mathcal{L}_{RLM}(\theta)$ from ablation studies with and without subtracting the encoder features of LM from the encoder features of HLM.
THR loss explains the mutual information $I_{\phi}(F_{<t}, G_{<t})$ between RLM and HLM, and the minimization of it regularizes indiscriminate text hallucination existing in RLM.
For the case `with subtracting LM', it shows that both $\mathcal{L}_{THR}(\theta,\phi)$ and $\mathcal{L}_{RLM}(\theta)$ decrease and converge according to the epoch.
However, for the case `without subtracting LM'\footnote{$\mathcal{L}_{THR}(\theta,\phi) = I_{\phi}(F_{<t}, F_{<t}^{\star})$ for THR loss without LM}, minimizing the $\mathcal{L}_{THR}(\theta,\phi)$ hinders the convergence of $\mathcal{L}_{RLM}(\theta)$.
This is because the encoder features that contribute a sentence are in both RLM and HLM, minimizing $\mathcal{L}_{THR}(\theta,\phi)$ without removing them from HLM becomes adversarial with learning from cross-entropy loss, which degrades the performance of the VGD system.
\begin{figure}[t]
	\centering
	\includegraphics[width=\linewidth]{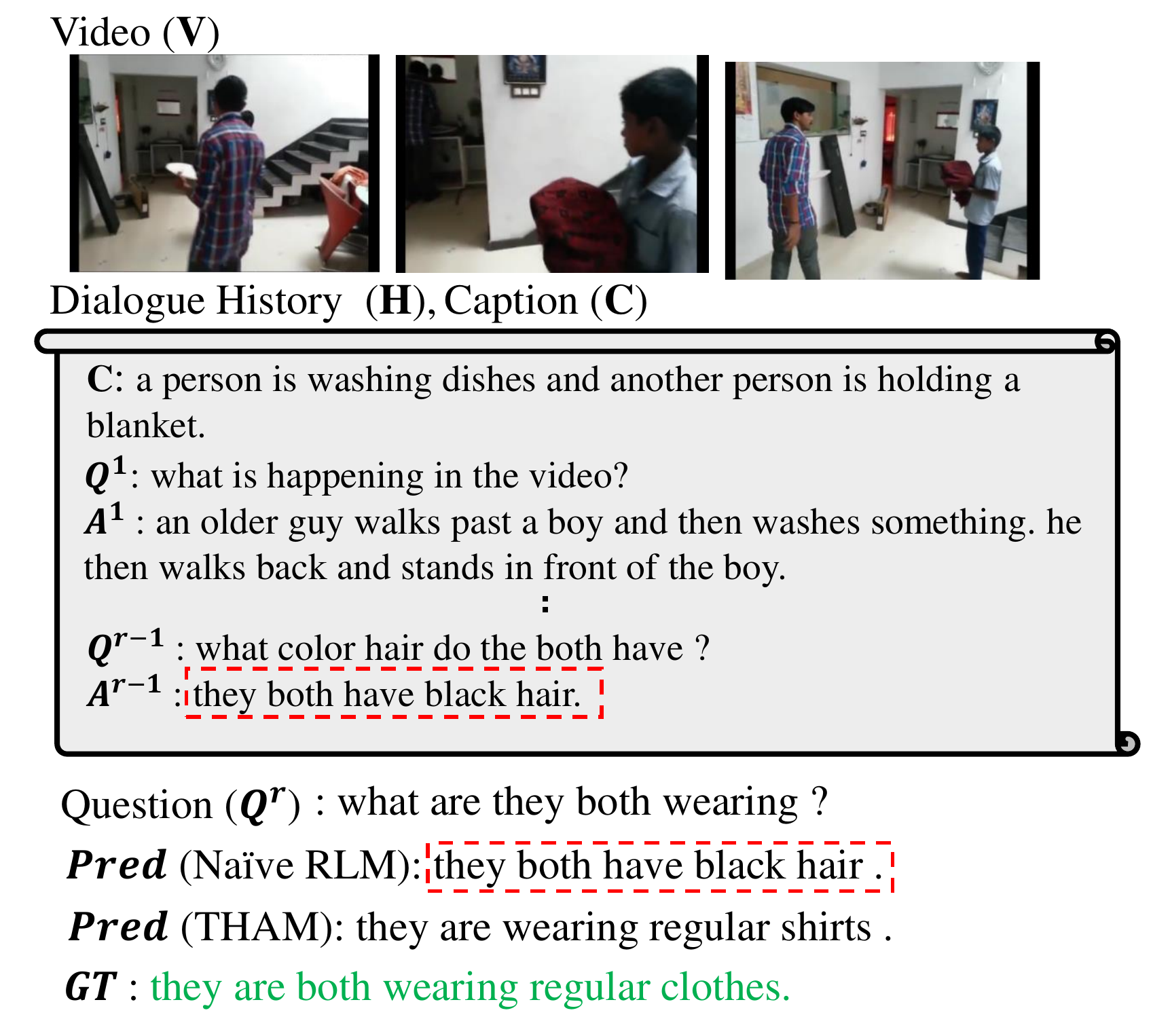}
	\caption{Response comparison between naive RLM and THAM on validation set of AVSD@DSTC7.}
	\label{fig:6}
\end{figure}
\subsection{Qualitative Results}
Figure \ref{fig:5} gives joint distributions among the language models' encoder features.
Here, the RLM is fully trained from THAM framework.
From 512 samples of AVSD validation set, we select a single value among the $d$-dimensional space at the same position of each encoder feature (\ie $F_{<t},F_{<t}^{\star},G_{<t}$).
Figure \ref{fig:5}(a) summarizes joint plots between $F_{<t}$ and $F_{<t}^{\star}$, where the correlations are confirmed due to the common grammatical knowledge from language models.
However Figure \ref{fig:5}(b) shows uncorrelated distributions between $F_{<t}$ and $G_{<t}$, which means the grammatical knowledge is properly removed from $G_{<t}$.
Figure \ref{fig:6} gives responses of naive RLM and THAM (naive RLM + THR loss).
For the question of ``what are they both wearing", naive RLM shows the reliance on texts from history without understanding of the question.
However, the THAM is generating correct answer sentence pertinent to the given question.
\section*{Conclusion}
Text Hallucination Mitigating framework is proposed for Video-grounded Dialogue.
THAM considers the text hallucination problem, which copies input texts for answer generation without understanding of the question.
THAM framework incorporates Text Hallucination Regularization loss derived from proposed information-theoretic text hallucination measurement approach.
Empirical results on VGD benchmarks show that THAM achieves state-of-the-art performances and effectiveness.
\section*{Acknowledgements}
This work was supported by Institute for Information \& communications Technology Promotion(IITP) grant funded by the Korea government(MSIT) (No. 2021-0-01381, Development of Causal AI through Video Understanding and Reinforcement Learning, and Its Applications to Real Environments) and partly supported by Institute of Information \& communications Technology Planning \& Evaluation (IITP) grant funded by the Korea government(MSIT) (No.2022-0-00184, Development and Study of AI Technologies to Inexpensively Conform to Evolving Policy on Ethics).
We would also like to thank the anonymous reviewers for their feedback. 
\section*{Limitations}
The limitations of the Text Hallucination Mitigating Framework are as follows.
First, our empirical analysis provides that THAM is facing a failure case about the question of sounds.
In Figure \ref{fig:7} in supplemental materials, for the question of ``what kind of noise", THAM is hallucinating response without understanding the question.
Although the answer ``i can hear some noise" can be plausible, but it also seems just hallucinating by copying from history texts. 
We speculate this is because the sound features contain less information ($128$ dimensions) comparing to video ($2048$ dimensions), which requires more specialized attention (\eg fine-grained audio processing).
For the second limitation, THAM is based on two-stage training mechanism.
To perform mitigation of text hallucination, pre-training of each language model is required as a first-stage training. 
To overcome the aforementioned limitations, we will perform further studies and make an effort on video interpretability improvements.

\section*{Ethics Statement}
As one of the interactive AI, the Video-grounded Dialogue system is designed for providing assistance to various subsections of our environments including education, entertainment, and visual impairments.
Our proposed Text Hallucination Mitigation Framework have contributed to improving response qualities and alleviating abnormalities in the system.
We also consider the potential negative societal impact that those who are aware of the VGD system can deliberately manipulate it to get prohibited information.
Furthermore, to apply the VGD system in the real environment, fairness and bias issues of dialogue systems should also be addressed.
%

\bibliography{anthology,custom}
\bibliographystyle{acl_natbib}

\clearpage
\appendix
\label{sec:appendix}
\section{Training Details.}
\paragraph{Training.} THAM is trained on NVIDIA TITAN V (12GB of memory) GPU with Adam optimizer \cite{kingma2014adam} with $\beta_{1}$ = 0.9, $\beta_{2}$ = 0.99, and $\epsilon$ = 10e-8. 
We utilized the piece-wise linearly decreased learning rate from 6.25e-4 to 0 and set the learning rate warm-up strategy to 10,000 training steps and trained the model up to 20 epochs. 
In Section 4.1, the interpolation is conducted via the window overlapping method.
The first-stage training is performed on three language models (\ie RLM, HLM, LM) respectively with a batch size of 8 and a dropout rate of 0.3.
For the $d$-dimensional space, all language models use $d$=768.
The second-stage training is performed on RLM with THR loss with the same batch size and dropout rate with the first training.
The best model is decided by the lowest validation loss on the validation-set with $\alpha=0.01$ in equation (11) of the main paper on the setting $X_{<t}^{\star}=[\mathbf{h}||\mathbf{a}_{<t}]$.
The training takes about 5 hours to be fully optimized at the losses of about 0.184 on training and 0.284 on validation.
Inference time for generating the answer for a single question takes about 2 seconds.
Our model is not performed on hyperparameter searching for model fine-tuning. 
\paragraph{Inference.} In the inference, answer generation adopts a beam search with a beam size of 5 and a length penalty of 1.0, where the maximum length of sentence is set to 30.
Every performance of THAM in table 1 and 2 of the main paper is averaging from 5 times random seed validation.

\section{Donsker-Varadhan representation.}
For the probability distribution of $P$ and $Q$, the KL divergence admits the following dual representation as:
\begin{eqnarray}
D_{KL}(P||Q) = \underset{T:\Omega \rightarrow \mathbb{R}}{sup} \mathbb{E}_{P}[T] - \textrm{log}(\mathbb{E}_{Q}[e^{T}]),
\end{eqnarray}
where $\Omega$ is high-dimensional variables and the supremum is taken over all functions $T$ such that the two expectiations are finite.
The proof for this representation is given as follows. For a given function $T$, consider the Gibbs distribution $G$ define by $dG = \frac{1}{z}e^{T}dQ$, where $Z = \mathbb{E}_{Q}[e^{T}]$. For the construction, we are available to derive\footnote{$\textrm{log}\frac{dG}{dQ} = \textrm{log}\frac{1}{Z}e^{T} = \textrm{log}\frac{1}{Z} + T = T - \textrm{log}Z$} as:
\begin{eqnarray}
\mathbb{E}_{P}[T] - \textrm{log}Z = \mathbb{E}_{P}[\textrm{log}\frac{dG}{dQ}]
\end{eqnarray}
Let $\Delta$ be the gap as:
\begin{eqnarray}
\Delta := D_{KL}(P||Q) - (\mathbb{E}_{P}[T] - \textrm{log}(\mathbb{E}_{Q}[e^{T}]))
\end{eqnarray}
Using the Equation (2), we can write $\Delta$ as KL-divergence:
\begin{align}
\begin{split}
\Delta = &\mathbb{E}_{P}[\textrm{log}\frac{dP}{dQ} - \textrm{log}\frac{dG}{dQ}]\\ &= \mathbb{E}_{P}\textrm{log}\frac{dP}{dG} = D_{KL}(P||G)
\end{split}
\end{align}
The positivity of the KL-divergence gives $\Delta \geq 0$. Therefore, we are able to show that for any $T$,
\begin{eqnarray}
D_{KL}(P||Q) \geq \mathbb{E}_{P}[T] - \textrm{log}(\mathbb{E}_{Q}[e^{T}]),
\end{eqnarray}
and the inequality is preserved via taking the supremum over the right-hand side, where the identity of Equation (4) also shows that the bound is tight whenever $G = P$, for optimal functions $T^{\star}$ taking the form $T^{\star} \ \textrm{log}\frac{dP}{dQ} + C$ for some constant $C \in \mathbb{R}$.
\begin{figure}[h!]
	\centering
	\includegraphics[width=\linewidth]{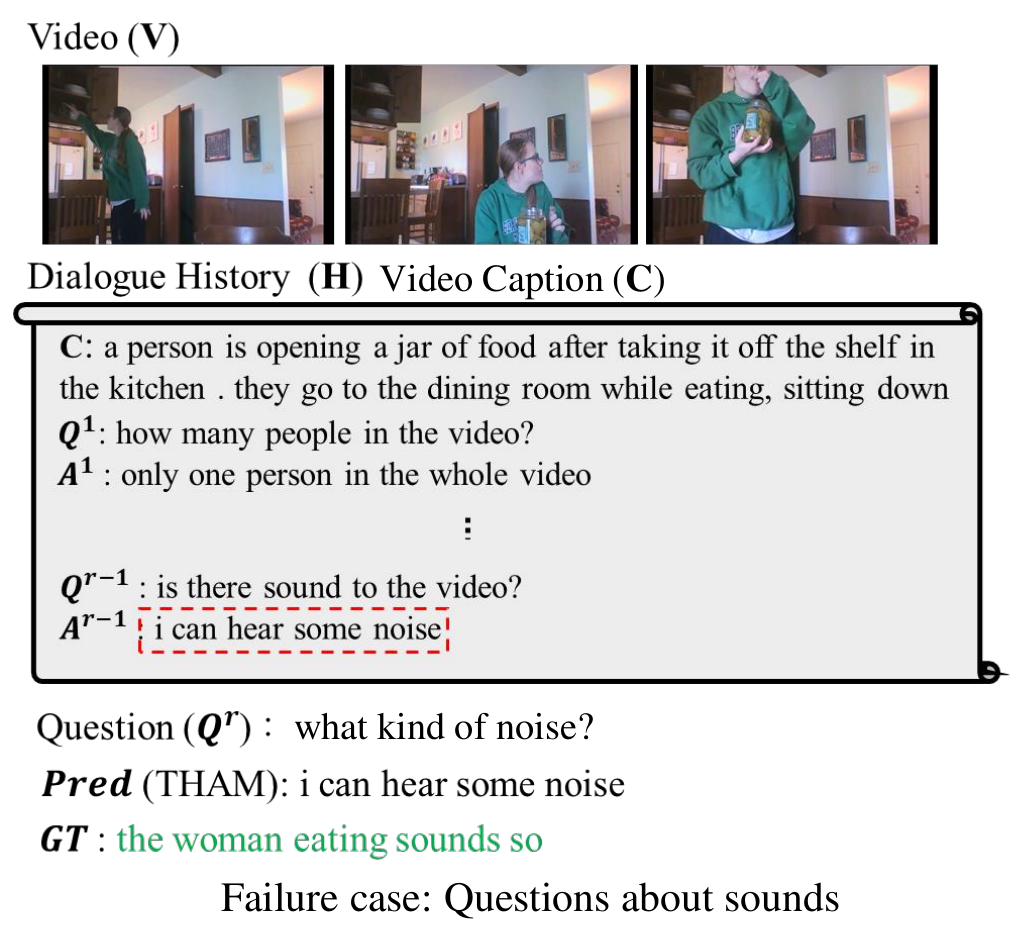}
	\caption{Failure case on question about sounds}
	\label{fig:7}
\end{figure}
\section{Failure case}
We also confirmed that the proposed THAM is fragile to the questions of asking sounds in the video, where it copies the input texts of ``i can hear some noise'' from history texts in Figure \ref{fig:7}.
While we admit that the above case can produce semantically correct answers, we feel that the VGD systems should be able to generate more rich answers using their own languages.
Furthermore, the sound features contain less information ($128$ dimensions) compared to video ($2048$ dimensions), which requires more specialized attention.
\end{document}